\documentclass[runningheads]{llncs}
\usepackage[T1]{fontenc}
\usepackage{xcolor}
\usepackage{graphicx}
\usepackage{multirow}
\usepackage{array}
\usepackage{graphicx,verbatim}
\usepackage{amsmath}
\usepackage{amssymb}
\usepackage{caption}
\usepackage[colorlinks=true,  
            linkcolor=blue,   
            citecolor=blue,   
            urlcolor=blue]{hyperref}
\usepackage[sort,compress]{cite}
\usepackage{booktabs}
\usepackage{multirow}
\usepackage{subcaption}

\newcommand{\modelacro}{{CALM }}
\newcommand{\modelacroNSP}{{CALM}}
\begin{document}
\title{CALM: Interpretable Cross-Modal Alignment for Biomarker Discovery from Unpaired Data}
\titlerunning{CALM: Interpretable Cross-Modal Alignment}
%
\author{Jueqi Wang\inst{1} \and
Zachary Jacokes\inst{2} \and
John Darrell Van Horn\inst{2,3} \and 
Kevin A. Pelphrey\inst{4} \and
Michael C. Schatz\inst{5} \and
Archana	Venkataraman\inst{1}
}


\authorrunning{J. Wang et al.}

\institute{Department of Electrical and Computer Engineering, Boston University \\
\email{\{jueqiw,archanav\}@bu.edu} \and
 School of Data Science, University of Virginia \\
\email{zj6nw@virginia.edu} \and
Department of Psychology, University of Virginia \\
\email{jdv7g@virginia.edu} \and
 Department of Neurology, University of Virginia \\
\email{kevin.pelphrey@virginia.edu} \and
 Departments of Computer Science and Biology, Johns Hopkins University \\
\email{mschatz@cs.jhu.edu}
}

  
\maketitle              
\begin{abstract}
The interaction between brain structure and genetic influences is key to understanding neuropsychiatric disorders. However, most large-scale datasets are unimodal, providing either neuroimaging or genetics data. We propose CALM, a framework that learns interpretable associations between brain ROIs and genetic pathways from \textit{completely disjoint} populations. CALM aligns the two modalities in a shared latent space via linear projections that simultaneously match the class-conditional latent distributions and ensure group separability. These projections provide interpretable pathway--ROI associations. When trained on unimodal imaging and genetics datasets, CALM generalizes to an unseen paired dataset, outperforming several state-of-the-art methods and ablation baselines. We also demonstrate stability of the learned associations against a paired baseline. Our experiments on autism spectrum disorder reveal immune and metabolic pathways linked to specific cortical regions and are consistent with established literature. Thus, CALM opens the door to leveraging large unimodal repositories for studying cross-modal interactions in brain disorders across disparate datasets.

\end{abstract}

\section{Introduction}

Disorders such as autism spectrum disorder (ASD) are driven by complex interactions between genetic risk factors and brain morphology~\cite{jack2021neurogenetic}. Understanding these relationships is crucial for biomarker identification; however, acquiring both neuroimaging and genetics data from the same individuals is costly and often impractical at scale. As a result, the largest public resources in precision psychiatry are non-overlapping and modality specific. For example, the Autism Brain Imaging Data Exchange (ABIDE)~\cite{di2014autism} offers comprehensive neuroimaging data, while the Simons Simplex Collection (SSC)~\cite{fischbach2010simons} provides extensive genomic data for ASD \textit{for a different set of participants}. Current multimodal analyses require paired samples
and cannot directly be applied to these large datasets.

Traditional imaging-genetics is based on 
linear models between a restricted set of neuroimaging and genetics features~\cite{shen2010whole}. This paradigm struggles 
to capture the complex dependencies associated with neuropsychiatric disorders~\cite{smedley2020discovering}. Deep learning offers a promising alternative for imaging-genetics. For instance, latent fusion methods combine low-dimensional encodings from each modality~\cite{ghosal2021g,10313442}; however, the information compression obscures the link between brain regions and genomic features, thus limiting biological interpretability. Early fusion approaches using cross-attention can explicitly model such interactions~\cite{wang2025learning,jaume2024modeling,smedley2020discovering}, but they cannot enforce consistency in learned associations across subjects, and they often require models with a large number of parameters~\cite{jaume2024modeling,wang2025learning}. Graph-based methods have been used to incorporate \textit{a priori} biological knowledge~\cite{ghosal2022biologically} but operate at the gene level and are prone to overfitting on small samples. Critically, these approaches require paired data from the same individuals, whereas public neuroimaging and genetic repositories are largely disjoint.

Recent works have addressed the problem of unpaired multimodal learning, but at the cost of interpretability~\cite{hoshen2018unsupervised}. For example, spectral embedding methods such as SUE~\cite{yacobi2025learning} align the modalities through a canonical correlation analysis and distribution matching, but the learned representations lack clear biological meaning. Additive fusion strategies require both modalities to share identical dimensionality, similarly obscuring interpretable structure~\cite{zhang2023learning,ghosal2021g}. In contrast, optimal transport (OT) methods compute a minimum-cost coupling between two distributions~\cite{cuturi2013sinkhorn,li2025alignmamba} and is the most natural when distributions describe comparable entities under a shared geometry. However, brain morphometry and genetic pathways are fundamentally different entities, and no biologically grounded metric exists to quantify distances between them, making the transport cost dependent on modeling assumptions rather than biological structure.


To address these limitations, we propose \underline{C}lass-conditional \underline{A}lignment with \underline{L}inear \underline{M}aps (\modelacroNSP), a framework that discovers interpretable associations between \textit{completely unpaired} neuroimaging and genetics datasets. \modelacro pretrains modality-specific encoders that preserve diagnostic groupings while ensuring a direct link between latent dimensions and neurobiological features. Cross-modal alignment is implemented as low-dimensional linear mappings in the latent space that optimize both a class-conditional alignment loss to match the latent imaging and genetics distributions within each diagnostic group, and a contrastive loss to pull same-class samples together while pushing apart different-class samples, regardless of modality. We validate \modelacro by training on ABIDE (neuroimaging) and SSC (genetics) and testing on the independent and paired ACE dataset. We show that associations recovered from unpaired data closely agree with those obtained from paired data and remain stable across cross-validation folds. \modelacro lays a crucial foundation to use independent unimodal datasets for cross-modal biomarker discovery, thus capitalizing on existing data resources. 

\begin{figure}[t]
    \includegraphics[width=\textwidth]{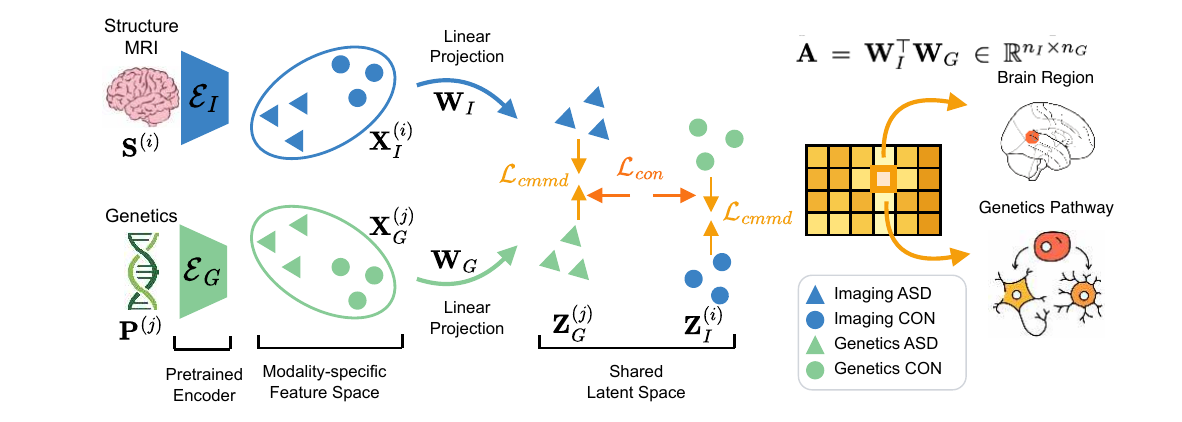}
    \caption{Overview \modelacroNSP. Given unpaired MRI and genetics data, pretrained encoders ($\mathcal{E}_I$, $\mathcal{E}_G$) output item-specific representations that are projected into a shared latent space via the learned linear maps $\mathbf{W}_I$ and $\mathbf{W}_G$. The loss function simultaneously matches the within-class distributions across modalities, while separating diagnostic groups.
    The cross-modal association matrix $\mathbf{A} = \mathbf{W}_I^\top \mathbf{W}_G$ quantifies the association between each brain region and genetic pathway.}
    \label{fig:ModelOverview}
\end{figure}

\section{Interpretable Linear Alignment for Unpaired Data}
\modelacro tackles the problem of discovering biomarker associations between neuroimaging and genetics data, when the modalities are collected from \textit{separate and non-overlapping groups of subjects}. Our framework is outlined in Fig.~\ref{fig:ModelOverview}.

The neuroimaging input to \modelacro consists of region-wise morphological features from structural MRI. Formally, let $\mathbf{S}^{(i)} \in \mathbb{R}^{n_I \times n_F}$ denote the neuroimaging input for subject~$i$, where $n_I$ is the number of brain regions, and $n_F$ is the number of morphological features. The genetics input to \modelacro is obtained by aggregating genetic risk scores across established biological pathways~\cite{kanehisa2000kegg}. Due to the heterogeneity of neuropsychiatric disorders, we include pathway-level features for several phenotypic traits. Details of this process are provided in Section~\ref{sec:2.5}. Formally, let $\mathbf{P}^{(j)} \in \mathbb{R}^{n_G \times n_T}$ denote the genetics input for subject~$j$, where $n_G$ and $n_T$ are the number of genetic pathways and phenotypic traits, respectively. 

\subsection{Data Encoding} 
\label{sec:dataencoding}
\modelacro uses pretrained autoencoders and classifiers to construct the latent representations for each modality. The loss combines reconstruction quality, classification accuracy, and a contrastive term that encourages latent representations from the same cohort to cluster together. This cohort-level grouping enables \modelacro to use linear alignment maps to discover stable correspondences between the two modalities. Crucially, the encoders consist of 
small per-entity layers that preserve the region-level information from the neuroimaging input and pathway-level information from the genetics input. This strategy 
preserves a one-to-one correspondence between rows and neurobiological entities, thus maintaining the interpretability of \modelacroNSP. Let $\mathbf{X}_I^{(i)} \in \mathbb{R}^{n_I \times d}$ be the latent neuroimaging representation for subject~$i$ and $\mathbf{X}_G^{(j)} \in \mathbb{R}^{n_G \times d}$ be the latent genetics representation for subject~$j$. The latent dimension~$d$ is shared between the modalities.

\subsection{Linear Projection} \label{sec:linproj}

Given the latent encodings $\mathbf{X}_I^{(i)}$ and $\mathbf{X}_G^{(j)}$, we map the modalities into a shared \\

\noindent latent space using learnable linear projections 
$\mathbf{W}_I \in \mathbb{R}^{d \times n_I}$ and 
$\mathbf{W}_G \in \mathbb{R}^{d \times n_G}$:
\begin{equation}
    \label{eq:encoding}
    \mathbf{Z}_I^{(i)} = \mathbf{W}_I \mathbf{X}_I^{(i)}, \quad 
    \mathbf{Z}_G^{(j)} = \mathbf{W}_G \mathbf{X}_G^{(j)}
\end{equation}

For convenience, we flatten the matrices in Eq.~\eqref{eq:encoding} into the vectors: 
$\mathbf{z}_I^{(i)} = \text{vec}(\mathbf{Z}_I^{(i)})$ and $\mathbf{z}_G^{(j)} = \text{vec}(\mathbf{Z}_G^{(j)})$ when defining the distributional loss in the next section. The association matrix $\mathbf{A} = \mathbf{W}_I^\top \mathbf{W}_G \in \mathbb{R}^{n_I 
\times n_G}$quantifies the similarity between cross-modal projection vectors, i.e., $\mathbf{A}_{r,p}=\mathbf{w}_{I,r}^\top\mathbf{w}_{G,p}$. As seen, $\mathbf{A}_{r,p}$ is large when ROI~$r$ and pathway~$p$ load along the same latent directions, indicating they co-vary in the aligned space. This formulation parallels that of canonical correlation analysis (CCA) but does not require paired samples.

\subsection{Class-Conditional Distribution Alignment}
\modelacro leverages a compound objective function to align the imaging and genetic distributions without requiring paired multi-modal samples:
\begin{equation}
    \mathcal{L}_{total} = \lambda_{cmmd} \mathcal{L}_{cmmd} + 
    \lambda_{con} \mathcal{L}_{con} + \lambda_{orth} \mathcal{L}_{orth}
\label{eq:total_loss}
\end{equation}
The loss terms in Eq.~\eqref{eq:total_loss} address three challenges of unpaired learning: stabilizing geometric alignment, preventing negative transfer between diagnostic groups, and enforcing semantic discriminability in the shared latent space. The hyperparameters $\lambda_{cmmd}, \lambda_{con}, \lambda_{orth}$ control the contribution of each term.

\medskip \noindent
\textbf{Distributional Loss ($\mathcal{L}_{cmmd}$):}
We use Class-Conditional Maximum Mean Discrepancy (CMMD) to align the distributions of latent neuroimaging and genetics encodings \textit{within each diagnostic group}. Unlike the standard MMD~\cite{gretton2012kernel}, which is label agnostic, $\mathcal{L}_{cmmd}$ calculates the discrepancy separately for each class~$c$: 
\begin{equation}
    \mathcal{L}_{cmmd} = \sum_{c \in \mathcal{C}} \left\| \frac{1}{N_c} 
    \sum_{i \in c} \phi(\mathbf{z}_I^{(i)}) - \frac{1}{M_c} 
    \sum_{j \in c} \phi(\mathbf{z}_G^{(j)}) \right\|_{\mathcal{H}}^2,
    \label{eq:cmmd}
\end{equation}
where $\phi(\cdot)$ maps features to a reproducing kernel Hilbert space (RKHS), and $N_c, M_c$ denote the number of imaging and genetic samples in class $c$, respectively.

In our experiments, the classes $c \in \mathcal{C}$ are ASD patients and neurotypical controls (NC), but the CMMD loss can be generalized to other scenarios. 

\medskip \noindent
\textbf{Contrastive Loss ($\mathcal{L}_{con}$):}
While $\mathcal{L}_{cmmd}$ aligns the within-class distributions, it does not enforce a separation \textit{between classes} in the shared latent space. Thus, we employ a supervised contrastive loss~\cite{khosla2020supervised} that treats all same-class samples as positives regardless of modality. Let $\mathcal{B}$ denote the set of latent representations from both modalities included within a mini-batch. For an anchor $\mathbf{z}^{(i)}$ with label $y_i \in \mathcal{C}$, the positive set contains all other samples sharing the same label, $\mathcal{P}(i) = \{j \neq i \in \mathcal{B} : y_j = y_i\}$. The contrastive loss is computed
\begin{equation}
    \mathcal{L}_{con} = - \sum_{i \in \mathcal{B}} 
    \frac{1}{|\mathcal{P}(i)|} \sum_{p \in \mathcal{P}(i)} 
    \log \frac{\exp(\mathrm{sim}(\mathbf{z}^{(i)}, 
    \mathbf{z}^{(p)}) / \tau)}{\sum_{j \in \mathcal{B} 
    \setminus \{i\}} \exp(\mathrm{sim}(\mathbf{z}^{(i)}, 
    \mathbf{z}^{(j)}) / \tau)},
    \label{eq:con}
\end{equation}
where $\tau$ is a temperature parameter, and $|\mathcal{P}(i)|$ is the total number of positive samples for anchor~$\mathbf{z}^{(i)}$. Eq.~\eqref{eq:con} simultaneously pulls samples with the same class together across modalities and pushes apart samples in different classes. This procedure encourages a shared discriminative structure in the latent space.

\medskip \noindent
\textbf{Regularization ($\mathcal{L}_{orth}$):}
To prevent degenerate solutions for the projection matrices, we use regularization to encourage orthonormality in the rows:
\begin{equation}
    \mathcal{L}_{orth} = \left\| \mathbf{W}_{I} \mathbf{W}_{I}^\top - \mathbf{I} \right\|_F^2 + \left\| \mathbf{W}_{G} \mathbf{W}_{G}^\top - \mathbf{I} \right\|_F^2
    \label{eq:orth}
\end{equation}
where $\|\cdot\|_F$ denotes the Frobenius norm, and $\mathbf{I}$ is the identity matrix. Each row of $\mathbf{W}_{I}$ and $\mathbf{W}_{G}$ defines the projection of a single latent dimension across input features. Encouraging their outer products to be identity ensures that the projections remain orthogonal with unit-norm. This prevents rank collapse and encourages each latent dimension to capture non-redundant information.

\subsection{Implementation Details}

\modelacro\footnote{Code is available at \url{https://github.com/jueqiw/CALM}} follows a two-stage training procedure: modality-specific encoders and classifiers are pretrained for 50 epochs, the classifiers are frozen while the linear projection matrices are trained for cross-modal alignment over 30 epochs. The inputs~$\mathbf{S^{(i)}}$ and~$\mathbf{P^{(j)}}$ are z-score normalized using per-feature statistics computed solely from the training set. The hyperparameters in Eq.~\eqref{eq:total_loss} were set to $\lambda_{cmmd} = 5$, $\lambda_{con} = 5$, and $\lambda_{orth} = 0.01$, with a contrastive temperature of $\tau = 0.07$. The projector learning rate was set to $1 \times 10^{-3}$, with a reduced rate of $1 \times 10^{-5}$ for fine-tuning the encoders. We used layer normalization, a batch size of 64, and applied Gaussian feature noise ($\sigma = 0.15$) to the imaging modality for regularization. The shared latent dimension was set to $d = 6$.


\medskip \noindent
\textbf{Class Prediction on Paired Test Data:} The distributional and contrastive losses refine both encoders during the second stage of training, so the updated representations now capture cross-modal structure. As a quantitative validation, we use the updated encoders and previously frozen stage one classifiers to predict the diagnosis of unseen test subjects. As described in Section~\ref{sec:2.5}, our test cohort contains paired data. Thus, the final prediction is obtained by averaging the output probabilities for both the neuroimaging and genetics classifiers.

\medskip \noindent
\textbf{Baseline Comparisons:} 
We evaluate the quantitative performance of \modelacro in predicting diagnostic labels against several state-of-the-art multimodal baselines. G-MIND~\cite{ghosal2021g} and UNSEEN~\cite{zhang2023learning} both project modality-specific latent encodings into a shared latent space and merge them through summation. Unlike \modelacroNSP, these methods blur information across input features, thus limiting our ability to extract feature-wise associations between the modalities. SUE~\cite{yacobi2025learning} aligns independently learned spectral embeddings across modalities without paired samples, but also lacks interpretable cross-modal associations. For consistency, all baselines use the same pathway inputs and imaging feature representations as \modelacroNSP. We also compare \modelacro against an optimal transport ablation that replaces the $\mathcal{L}_{cmmd}$ loss with a global Sinkhorn divergence~\cite{cuturi2013sinkhorn}. Finally, we include two other ablations of \modelacroNSP, omitting the distributional loss $\mathcal{L}_{cmmd}$ and omitting the contrastive loss $\mathcal{L}_{con}$, to assess their contributions.


\section{Data and Preprocessing} \label{sec:2.5}

We evaluate CALM using four publicly available datasets for ASD. For training, we use neuroimaging data from the ABIDE~I~and~II datasets~\cite{di2014autism} ($N=1,898$; $889$ ASD) and genetics data from SSC~\cite{fischbach2010simons} ($N=2,452$). For SSC simplex families, we use proband genotypes as ASD samples and construct pseudo-controls from untransmitted parental alleles~\cite{falk1987haplotype}.
For testing, we use the \textit{independent} ACE\footnote{Data available from \url{https://nda.nih.gov/edit_collection.html?id=2021}} dataset ($N=198$; $105$ ASD), which contains paired MRI and genetic data. 

\medskip \noindent
\textbf{Neuroimaging:} T1-weighted MRI from ABIDE and ACE were preprocessed using FreeSurfer~\cite{fischl2012freesurfer} and parcellated according to the Brainnetome atlas~\cite{li2023brainnetome} ($n_{I} = 246$ ROIs). For each ROI, we extract $n_{F}=4$ structural features: volume size, surface area, average cortical thickness, and standard deviation of the cortical thickness. 
We use ComBat~\cite{johnson2007adjusting} to remove ABIDE site effects. The input features~$\mathbf{S}^{(i)}$ were normalized based on statistics from the training set.

\begin{table}[t]
    \setlength{\tabcolsep}{5pt}
    \centering
    \caption{Classification performance on ACE. Models are trained on ABIDE (neuroimaging) and SSC (genetics). The best performance is highlighted in bold. * denotes statistically worse performance than the best model (p < 0.05).}
    \label{tab:quantitative_results}
    {\begin{tabular}{l|cccc}
        \toprule
        Methods & Accuracy $\uparrow$ & Sensitivity $\uparrow$ & Specificity $\uparrow$ & AUC $\uparrow$ \\
        \midrule
        G-MIND~\cite{ghosal2021g} & 0.518$\pm$0.01* & 0.533$\pm$0.04* & 0.490$\pm$0.04* & 0.540$\pm$0.01* \\
        UNSEEN~\cite{zhang2023learning} & 0.499$\pm$0.02* & 0.490$\pm$0.25 & 0.510$\pm$0.23 & 0.522$\pm$0.02*  \\
        SUE~\cite{yacobi2025learning} & 0.486$\pm$0.05* & 0.430$\pm$0.11* & 0.549$\pm$0.09 & 0.478$\pm$0.04* \\
        Ablation \(\mathcal{L}_{OT}\)~\cite{cuturi2013sinkhorn} & 0.567$\pm$0.03 & 0.573$\pm$0.08 & \textbf{0.568$\pm$0.06} & 0.599$\pm$0.04 \\
        Ablation \(-\mathcal{L}_{cmmd}\) & 0.570$\pm$0.03 & 0.568$\pm$0.08 & 0.548$\pm$0.08 & 0.585$\pm$0.04 \\
        Ablation \(-\mathcal{L}_{con}\) & 0.568$\pm$0.03 & 0.575$\pm$0.09 & 0.557$\pm$0.09 & 0.598$\pm$0.04 \\
        \modelacroNSP & \textbf{0.572$\pm$0.03} & \textbf{0.583$\pm$0.08} & 0.562$\pm$0.06 & \textbf{0.606$\pm$0.03}  \\
        \bottomrule
    \end{tabular}}
\end{table}

\medskip \noindent
\textbf{Genetics:} SSC and ACE participants were genotyped on different Illumina chips.
Quality control was performed separately per batch and cohort using PLINK~\cite{purcell2007plink}. Analyses were restricted to a European subpopulation identified via projection onto the first three principal components of the 1,000 Genomes reference panel~\cite{10002015global}. 
We used the genetics preprocessing steps outlined in~\cite{marees2018tutorial}, which after LD clumping, yielded 44,102~SNPs. 
The SNPs were grouped into 3,907 genes defined by the $n_P = 177$ non-cancer pathways in the KEGG database~\cite{kanehisa2000kegg}.

Following~\cite{wang2025learning}, each entry $\mathbf{P}^{(n)}_{gt}$ in the input $\mathbf{P}^{(n)} \in \mathbb{R}^{n_G \times n_T}$ corresponds to the genetic risk for a single pathway~$g$ and phenotypic trait~$t$. This value is obtained by first selecting the SNPs associated with the pathway genes, weighting them by the effect sizes obtained from a published GWAS~\cite{wightman2021genome}, and computing a weighted sum across SNPs. We capture genetic risk across multiple phenotypes by leveraging the effect sizes from $m=6$ independent GWAS spanning disorder-related traits: ASD~\cite{pedersen2018ipsych2012,zhou2022integrating}, ADHD~\cite{demontis2023genome}, educational attainment (EA)~\cite{okbay2022polygenic}, intelligence~\cite{savage2018genome}, major depression disorder (MDD)~\cite{wray2018genome}, and epilepsy~\cite{international2023gwas}.

\begin{figure}[t]
    \centering
    \includegraphics[width=0.9\textwidth]{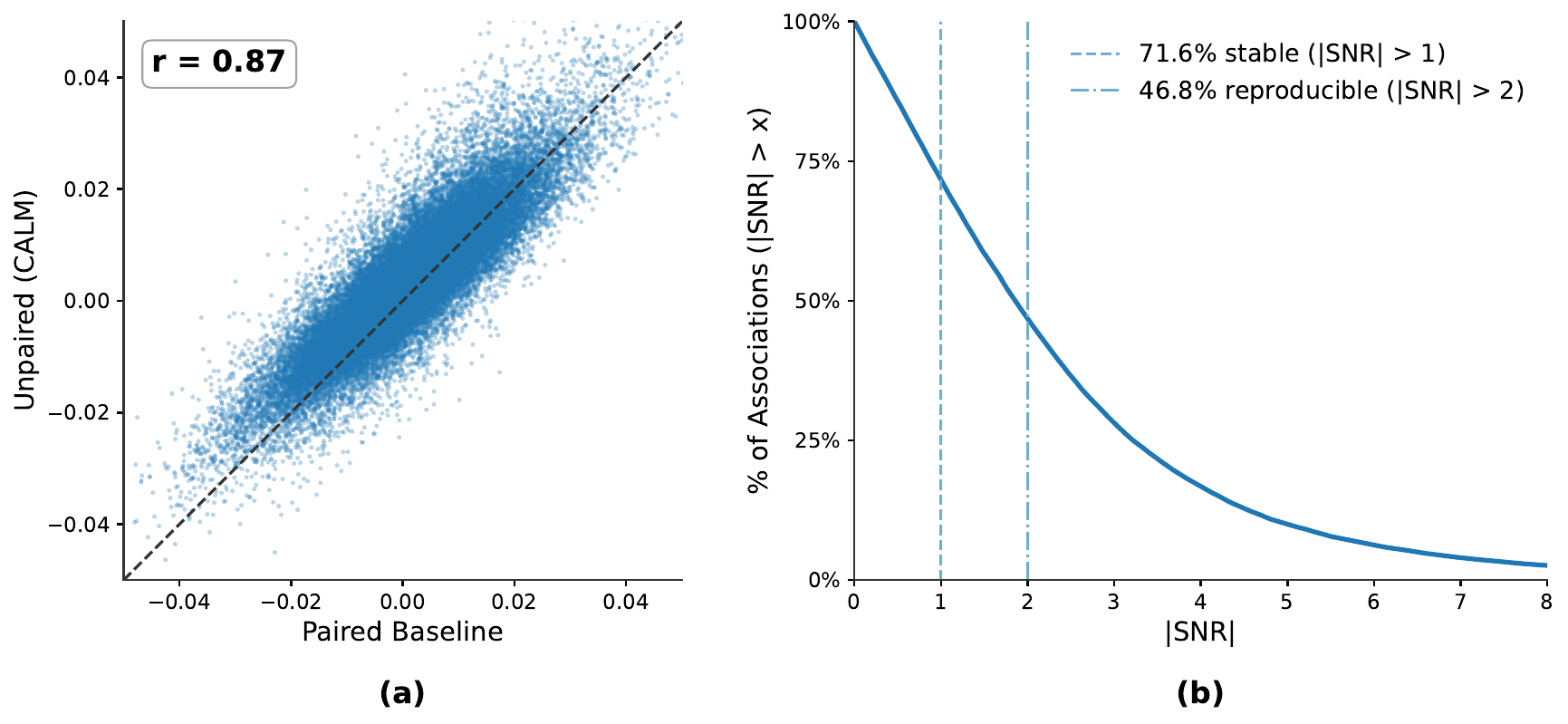}
    \caption{Validation of learned pathway-ROI associations. (a) Correlation between elements of the mean association matrix $\bar{\mathbf{A}}$ learned by \modelacro using unpaired data and when learned using a paired baseline. (b) Cross-fold stability measured by the survival function of signal-to-noise ratio (SNR) across five folds.}
    \label{fig:robust}
\end{figure}

\section{Experimental Results}
\textbf{Quantitative Disease Prediction:} Table~\ref{tab:quantitative_results} reports the ASD vs. NC classification performance on the ACE dataset when the models are trained on ABIDE and SSC within a 5-fold CV setup. As seen, \modelacro achieves the best performance across most metrics, with statistically significant improvement over G-MIND, UNSEEN, and SUE. 
While below the significance threshold, we note the consistent improvement of \modelacro over its model ablations. This improvement can be attributed to the complementary roles of the two loss terms. 

\begin{figure}[t]
    \includegraphics[width=0.95\textwidth]{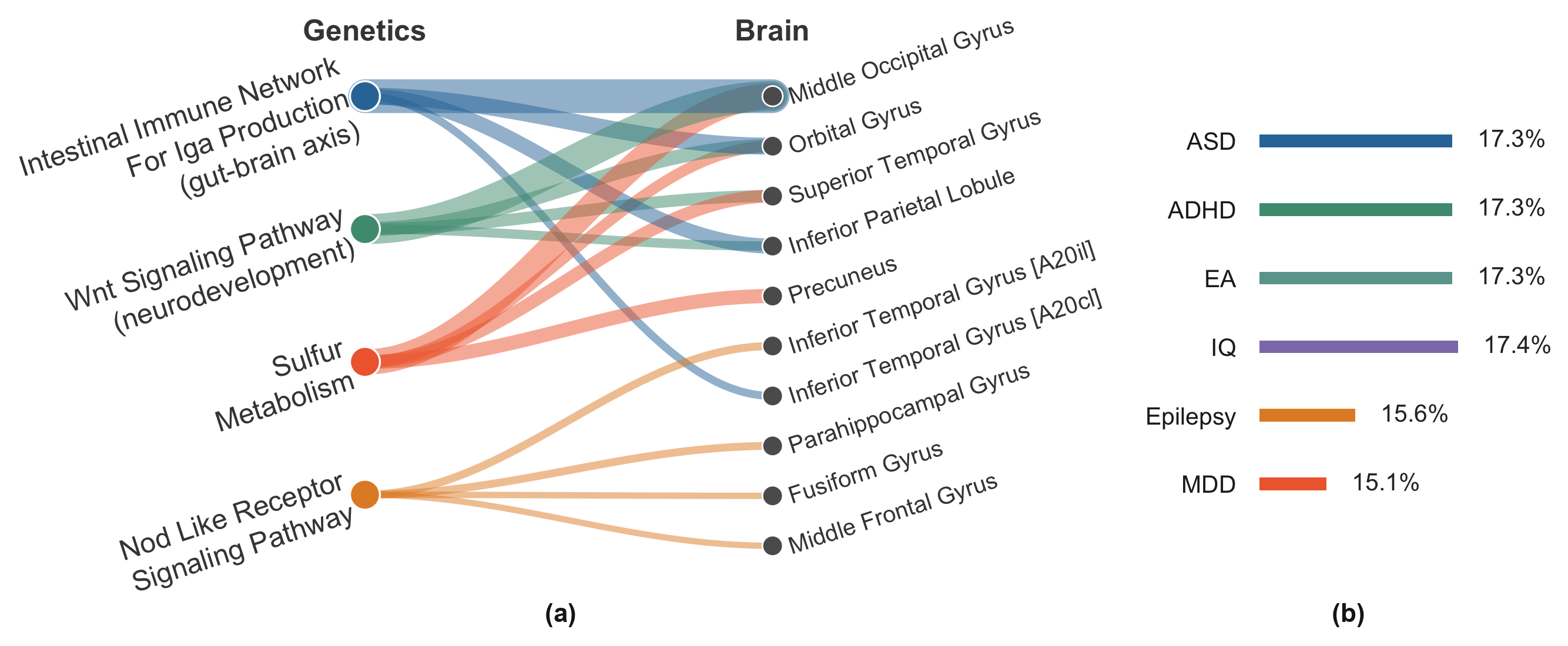}
    \caption{(a) Bipartite graph showing the top pathway-ROI associations; edge thickness encodes association strength. (b) Gradient-based attribution of each GWAS disorder channel to the shared latent space, normalized to sum to 100\%.}
    \label{fig:analysis}
\end{figure}

\medskip \noindent  
\textbf{Association Validation:} To validate that \modelacro learns pathway--ROI associations consistent with those obtained from true paired data, we replace the class-conditional distribution matching loss ($\mathcal{L}_{\text{CMMD}}$) with a paired contrastive loss (InfoNCE~\cite{oord2018representation}) that directly maps modalities from the same subject to similar latent representations. Fig.~\ref{fig:robust}(a) correlates the entries of the mean association matrix $\bar{\mathbf{A}}$ between the unpaired \modelacro and the paired baseline. The strong Pearson correlation ($r = 0.87$) indicates that CALM recovers consistent pathway--ROI associations without requiring paired data. We also assess cross-fold stability within the unpaired setting by computing the signal-to-noise ratio ($\text{SNR} = \mu / \sigma$) of each association across five folds. Fig.~\ref{fig:robust}(b) shows that 72\% of associations achieve $|\text{SNR}| > 1$ (stable) and 47\% achieve $|\text{SNR}| > 2$ (reproducible), confirming that most associations are robust across data splits.

\medskip \noindent  
\textbf{Learned Imaging-Genetics Associations:}
To determine the most salient pathway--brain ROI influences, we first rank pathways by their $\ell_2$ column norm of the association matrix $\mathbf{A}$
and select the top four pathways with the strongest brain-wide effects. We then merge bilateral ROIs by summing the association values of their left and right counterparts. For visualization, we display the top four brain ROIs with the highest association strength for each pathway in Fig.~\ref{fig:analysis}.

The IgA Immune Network, Wnt Signaling, and Sulfur Metabolism pathways converge on occipital, temporal, and parietal regions, linked to atypical visual and multisensory processing in ASD~\cite{nickl2012brain}. These pathways are functionally related through homeostatic regulation of neurodevelopment mechanisms that are well-established in ASD pathology~\cite{pelphrey2011research}. In contrast, NOD-like Receptor Signaling maps onto different ventral temporal and limbic areas, aligning with socioemotional and face processing deficits in ASD~\cite{hughes2018immune}.


\medskip \noindent
\textbf{Channel Attribution Analysis:} To quantify the contribution of each phenotypic trait to the shared latent space, we compute gradient-based input attribution~\cite{shrikumar2016not} through the genetics encoder $\mathcal{E}_G$ and the learned projection matrix $\mathbf{W}_G$, averaged across the top four pathways and five cross-validation folds. \textcolor{blue}{Fig. \ref{fig:analysis}(b)} shows that ASD, ADHD, EA, and IQ contribute most strongly, while MDD and epilepsy contribute moderately. This is consistent with the well-documented genetic and phenotypic overlap among these neurodevelopmental traits~\cite{litman2025decomposition}.

\section{Conclusion}  
We have proposed CALM, a framework that learns interpretable cross-modal associations between brain ROI morphology and genetic pathways from \textit{entirely unpaired and disjoint} populations. The backbone of \modelacro is a class-conditional linear transport model that aligns the group-wise distributions across modalities in the a shared latent space. We validate the learned associations by comparing them against a paired baseline trained on matched subjects, and we demonstrate the stability of the association matrix. CALM generalizes to a completely unseen external dataset and surpasses state-of-the-art baselines. The discovered pathway-ROI associations are supported by established ASD literature.

\begin{credits}
\subsubsection{\ackname} This work was supported by the National Institutes of Health awards 1R01HD108790 (PI Venkataraman).

\end{credits}

\bibliographystyle{splncs04}
\bibliography{Paper-2961}
\end{document}